\documentclass{article}

\usepackage[main, preprint]{mystyle}

\usepackage[utf8]{inputenc} 
\usepackage[T1]{fontenc}    
\usepackage{natbib}
\usepackage{url}            
\usepackage{booktabs}       
\usepackage{amsfonts}       
\usepackage{nicefrac}       
\usepackage{microtype}      
\usepackage{xcolor}         
\usepackage{subcaption}
\usepackage{setspace}
\usepackage{graphicx}
\usepackage{multirow}
\usepackage{amsmath}
\usepackage{adjustbox}
\usepackage{caption}
\usepackage{enumitem}
\usepackage{makecell}
\usepackage{colortbl}
\usepackage{wrapfig}
\usepackage{pifont}
\usepackage[most]{tcolorbox}
\usepackage{algorithm}          
\usepackage{listings}           
\usepackage{marvosym}
\usepackage{bxcoloremoji}
\setlist[itemize]{leftmargin=*}
\definecolor{mydarkblue}{rgb}{0,0.08,0.45}

\usepackage{amsmath,amsfonts,bm}

\def\eqref#1{equation~\ref{#1}}

\def\1{\bm{1}}

\DeclareMathAlphabet{\mathsfit}{\encodingdefault}{\sfdefault}{m}{sl}
\SetMathAlphabet{\mathsfit}{bold}{\encodingdefault}{\sfdefault}{bx}{n}

\usepackage{graphicx}
\usepackage{booktabs}
\usepackage{amssymb}
\usepackage{stmaryrd}
\usepackage[table]{xcolor}
\usepackage{pifont}
\usepackage{subcaption}

\definecolor{mygray}{HTML}{808080}
\definecolor{mydarkgreen}{HTML}{0BDA51}

\usepackage{listings}
\usepackage{hyperref}
\usepackage{url}
\usepackage{amsmath,amssymb,mathtools}
\usepackage{amsthm}
\usepackage{enumitem}
\usepackage{booktabs}
\usepackage{algorithm}
\usepackage{algpseudocode} 

\usepackage[table]{xcolor}
\usepackage{array}
\newcolumntype{g}{>{\columncolor{gray!10}}c}
\usepackage{multirow}           
\usepackage{adjustbox}
\usepackage{hyperref}
\usepackage{url}
\usepackage{subcaption}
\usepackage{booktabs} 
\usepackage{array}
\usepackage{algorithm}
\usepackage{algpseudocode} 
\usepackage{amsmath}
\definecolor{step1color}{HTML}{D5E8D4} 
\definecolor{step2color}{HTML}{A8D08D} 
\definecolor{step3color}{HTML}{82B366} 
\definecolor{step4color}{HTML}{5A8A42} 
\definecolor{promptcolor}{HTML}{F5F5F5} 
\definecolor{cond}{HTML}{2E75B6}

\title{{\color{cond} MM-CondChain}: A Programmatically Verified Benchmark for Visually Grounded\\ Deep Compositional Reasoning}

\author{
    \textbf{Haozhan Shen}$^{1,2}$\thanks{This work was done during an internship at Accio Team, Alibaba Group.}\quad
    \textbf{Shilin Yan}$^{1\dagger}$ \quad
    \textbf{Hongwei Xue}$^{1}$\textsuperscript{\ddag} \quad
    \textbf{Shuaiqi Lu}$^{1}$\quad
    \textbf{Xiaojun Tang}$^{1}$ \\[2pt]
    \textbf{Guannan Zhang}$^{1}$\quad
    \textbf{Tiancheng Zhao}$^{3}$\textsuperscript{\ddag}\quad
    \textbf{Jianwei Yin}$^{2}$\quad 
    \and \\
    $^{1}$Accio Team, Alibaba Group\quad    $^{2}$Zhejiang University\quad   
    $^{3}$ZJU-BJ \\ [2pt]
    \texttt{\{tattoo.ysl,xuehongwe\}@gmail.com}
    \and \\
    $^{\dagger}$ Project Leader \quad \textsuperscript{\ddag} Corresponding Author
}

\headerlogos{
	\includegraphics[height=0.5cm]{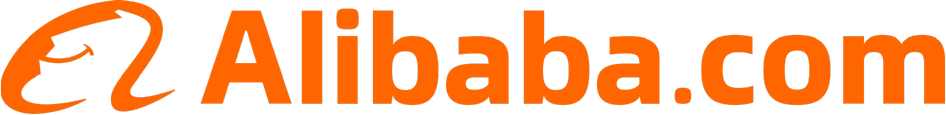}\hspace{0.3cm}
	\includegraphics[height=0.5cm]{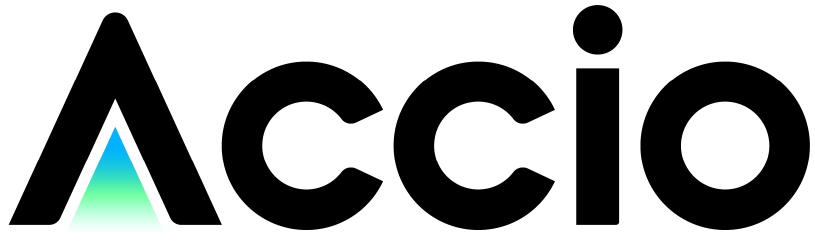}  
}

\newcommand{\github}{\raisebox{-1.5pt}{\includegraphics[height=1.05em]{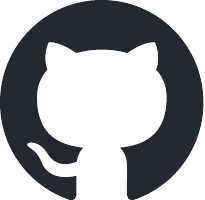}}}
\newcommand{\LSPbox}[2]{%
  \ifcase#1\relax
    \colorbox{white}{#2}%
  \or
    \colorbox{step1color}{\strut #2}%
  \or
    \colorbox{step2color}{\strut #2}%
  \or
    \colorbox{step3color}{\strut #2}%
  \or
    \colorbox{step4color}{\strut #2}%
  \else
    \colorbox{gray!20}{\strut #2}%
  \fi
}
\begin{document}
	
\maketitle
\vspace{-0.5cm}
\begin{abstract}
Multimodal Large Language Models~(MLLMs) are increasingly used to carry out visual workflows such as navigating GUIs, where the next step depends on verified visual compositional conditions (e.g., ``if a permission dialog appears and the color of the interface is green, click Allow'') and the process may branch or terminate early.
Yet this capability remains under-evaluated: existing benchmarks focus on shallow-compositions or independent-constraints rather than deeply chained compositional conditionals.
In this paper, we introduce \textbf{{\color{cond} MM-CondChain}}, a benchmark for \emph{visually grounded deep compositional reasoning}.
Each benchmark instance is organized as a multi-layer reasoning chain, where every layer contains a non-trivial compositional condition grounded in visual evidence and built from multiple objects, attributes, or relations.
To answer correctly, an MLLM must perceive the image in detail, reason over multiple visual elements at each step, and follow the resulting execution path to the final outcome.
To scalably construct such workflow-style data, we propose an agentic synthesis pipeline: a Planner orchestrates layer-by-layer generation of compositional conditions, while a Verifiable Programmatic Intermediate Representation (VPIR) ensures each layer's condition is mechanically verifiable. A Composer then assembles these verified layers into complete instructions.
Using this pipeline, we construct benchmarks across three visual domains: natural images, data charts, and GUI trajectories.
Experiments on a range of MLLMs show that even the strongest model attains only 53.33 Path F1, with sharp drops on hard negatives and as depth or predicate complexity grows, confirming that deep compositional reasoning remains a fundamental challenge. \\

\coloremojicode{1F310}  \textbf{Project Page:} \href{https://accio-lab.github.io/MM-CondChain}{https://accio-lab.github.io/MM-CondChain} \vspace{0.1cm}

\github{}  \textbf{Github Repo:} \href{https://github.com/Accio-Lab/MM-CondChain}{https://github.com/Accio-Lab/MM-CondChain} \vspace{0.1cm}

\coloremojicode{1F917}  \textbf{HuggingFace:} \href{https://huggingface.co/datasets/Accio-Lab/MM-CondChain}{https://huggingface.co/datasets/Accio-Lab/MM-CondChain}

\end{abstract}

\section{Introduction}
\label{sec:intro}

As Large Language Models~(LLMs)~\cite{abdin2024phi,achiam2023gpt4,anthropic_claude_opus_46_system_card_2026,yang2025qwen3,jiang2025d,qwen3.5,deepmind_gemini3pro_model_card,li2025adaptive,grattafiori2024llama3,liu2024deepseek-v3} and Multimodal Large Language Models~(MLLMs) ~\cite{achiam2023gpt4,openai_gpt5_system_card,deepmind_gemini3pro_model_card,deepmind_gemini3flash_model_card,deepmind_gemini25pro_modelcard_pdf,deepmind_gemini25flash_modelcard_pdf,qwen3.5,bai2025qwen3vl,yan2025crosslmm,anthropic_claude_opus_46_system_card_2026,hong2025glm-v} grow more capable, they are increasingly expected to go beyond simple visual question answering and tackle complex visual workflows where the correct action depends on a chain of visual checks (e.g., \emph{if a dialog appears, verify it requests location access; if so and the app is trusted, click Allow; otherwise...}).
These tasks require visually grounded deep compositional reasoning: at each step, the model must verify a multi-factor visual condition, and then determines whether the workflow continues or terminates early.
Thus, a natural question arises: \textbf{can current advanced MLLMs reliably follow deeply compositional condition instructions that require verification against visual input at every step?}

\begin{figure}[t]
  \centering
  \includegraphics[width=\linewidth]{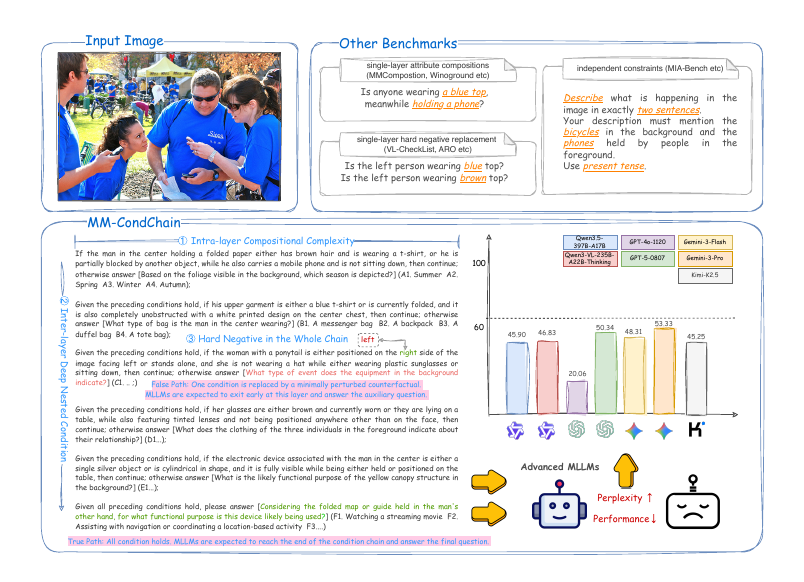}
  \caption{\textbf{MM-CondChain targets visually grounded deep conditional reasoning beyond prior benchmarks.}
\textit{Top}: existing benchmarks typically evaluate either shallow single-layer visual compositions or independent instruction constraints.
\textit{Bottom left}: \textbf{MM-CondChain} introduces nested, inter-layer conditional chains with rich intra-layer compositional predicates, where a minimally perturbed condition can create a hard negative that changes the execution path and causes early termination.
\textit{Bottom right}: experiments show that even advanced MLLMs achieve limited performance on this benchmark, highlighting visually grounded deep compositional reasoning as a fundamental challenge.}
  \label{fig:overview}
\end{figure}

Answering this question requires a benchmark that systematically probes such capabilities. However, existing benchmarks fall short in two key respects.
\emph{First}, in compositional depth. Prior visual reasoning benchmarks~\cite{hsieh2023sugarcrepe,johnson2017clevr,hudson2019gqa,hua2024mmcomposition} typically evaluate single-layer compositions (e.g., ``Is the object red and large?''), while instruction-following benchmarks~\cite{ifeval,jiang-etal-2024-followbench,qian2024mia,wen2024benchmarking,ifbench,Mm-ifengine} focus on independent constraints. Neither requires models to perform \emph{deep compositional reasoning} across layers. In these tasks, the model must verify a multi-factor visual condition at each step, and the outcome of each step then determines the subsequent reasoning path.
\emph{Second}, in the difficulty of hard negatives. Some prior benchmarks include contrastive pairs for compositional understanding~\cite{thrush2022winogroundprobingvisionlanguage,ARO,https://doi.org/10.48550/arxiv.2207.00221,zhao2022explainable}, but these are usually limited to a single-layer change, such as replacing one attribute or relation.

%

\newcommand{\cmark}{\textcolor{mydarkgreen}{\ding{51}}}
\newcommand{\xmark}{\textcolor{mygray}{\ding{55}}}

\begin{table}[t]
\centering
\caption{\textbf{Comparison with existing benchmarks.} \textbf{Compose}: intra-layer multi-attribute composition; \textbf{Nested}: inter-layer chained conditions; \textbf{Visual}: conditions grounded in visual input; \textbf{Hard Neg.}: contrastive pairs with minimal perturbation; \textbf{Prog. Verif.}: ground truth verified via code execution; \textbf{Determ.}: deterministic evaluation without LLM-as-judge; \textbf{Auto.}: automated data construction.}
\label{tab:comparison}
\small
\setlength{\tabcolsep}{6pt}
\resizebox{\columnwidth}{!}{%
\begin{tabular}{l|ccccccc}
\toprule
\textbf{Benchmark} & \textbf{Compose} & \textbf{Nested} & \textbf{Visual} & \textbf{Hard Neg.} & \textbf{Prog. Verif.} & \textbf{Determ.} & \textbf{Auto.} \\
\midrule
\multicolumn{8}{l}{\textbf{\textit{Visual Reasoning}}} \\
SugarCrepe~\cite{hsieh2023sugarcrepe} & \cmark & \xmark & \cmark & \cmark & \xmark & \cmark & \cmark \\
Winoground~\cite{thrush2022winogroundprobingvisionlanguage} & \cmark & \xmark & \cmark & \cmark & \xmark & \cmark & \xmark \\
ARO~\cite{ARO} & \cmark & \xmark & \cmark & \cmark & \xmark & \cmark & \xmark \\
MMComposition~\cite{hua2024mmcomposition} & \cmark & \xmark & \cmark & \xmark & \xmark & \cmark & \xmark \\
\midrule
\multicolumn{8}{l}{\textbf{\textit{Instruction Following}}} \\
IFEval~\cite{ifeval} & \xmark & \xmark & \xmark & \xmark & \cmark & \cmark & \xmark \\
FollowBench~\cite{jiang-etal-2024-followbench} & \xmark & \xmark & \xmark & \xmark & \xmark & \xmark & \xmark \\
MIA-Bench~\cite{qian2024mia} & \cmark & \xmark & \cmark & \xmark & \xmark & \xmark & \xmark \\
ComplexBench~\cite{wen2024benchmarking} & \cmark & \cmark & \xmark & \xmark & \xmark & \xmark & \xmark \\
MM-IFEval~\cite{Mm-ifengine} & \cmark & \xmark & \cmark & \xmark & \xmark & \xmark & \cmark \\
\midrule
\textbf{MM-CondChain} & \cmark & \cmark & \cmark & \cmark & \cmark & \cmark & \cmark \\
\bottomrule
\end{tabular}%
}
\end{table}

To address these gaps, we introduce \textbf{{\color{cond} MM-CondChain}}, a benchmark for \emph{visually grounded deep compositional reasoning} in MLLMs. 
Unlike prior benchmarks that test shallow-compositions or independent-constraints, \textbf{{\color{cond} MM-CondChain}} requires models to follow \emph{multi-layer control flow} where each decision is gated by a compositional condition that must be verified against the visual input, and where the execution may branch or terminate early.

However, building this kind of benchmark at scale is challenging.
If we directly ask an MLLM agent to generate long, multi-layer visual reasoning chains, the results often contain logical conflicts, unclear visual references, or statements that cannot be reliably determined from the visual input.
To address this, we decouple logical construction from natural-language writing through the proposed Verifiable Programmatic Intermediate Representation (VPIR).
Instead of generating the final instruction directly, we first represent each layer as an executable, Python-like predicate and mechanically verify whether it is true or false against structured visual facts, and only then translate the verified logic into natural language.
This makes the benchmark construction process reliable, controllable, and grounded in verifiable visual evidence.

Building on VPIR, we further develop an agentic synthesis pipeline that incrementally constructs each benchmark instance, as illustrated in Figure~\ref{fig:overview}.
At each layer, the pipeline generates a visually grounded compositional condition, verifies it mechanically against structured visual facts, and only then extends the reasoning chain.
VPIR explicitly represents both the verified condition and its minimally perturbed counterfactual at each layer, which naturally enables chained hard negatives.
As shown in Figure~\ref{fig:overview}, flipping a single predicate can change the execution path while keeping the overall instruction nearly unchanged, thereby forcing the model to accurately verify every condition along the way.
Compared with prior benchmarks, which mainly test shallow compositions or independent constraints, our benchmark targets deep, multi-layer reasoning with chained hard negatives.
Table~\ref{tab:comparison} summarizes the differences between \textbf{{\color{cond} MM-CondChain}} and existing benchmarks.

Using this pipeline, we instantiate \textbf{{\color{cond} MM-CondChain}} across three visual domains: natural images, data charts, and GUI trajectories.
Experiments on a range of state-of-the-art MLLMs show that visually grounded deep compositional reasoning remains highly challenging: even the strongest model achieves only 53.33 average Path F1, performance drops sharply on False-path hard negatives, and accuracy further degrades as reasoning depth and predicate complexity increase.

Our contributions are summarized as follows:
\begin{itemize}
    \item We introduce \textbf{{\color{cond} MM-CondChain}}, the first benchmark for visually grounded deep compositional reasoning, featuring multi-layer control flow with chained hard negatives.
    \item We propose a VPIR-based agentic synthesis pipeline that decouples logical construction from language rendering, enabling scalable benchmark construction with mechanical verifiability.
    \item We instantiate the framework across three visual domains and evaluate ten MLLMs, showing that even state-of-the-art models struggle with fine-grained verification of compositional visual conditions, especially on hard-negative instances and under greater depth or predicate complexity.
\end{itemize}
\section{Related Work}

\paragraph{Programmatically Verifiable Evaluation.}

IFEval~\cite{ifeval} introduced verifiable instructions whose compliance can be checked by simple Python functions, focusing on surface-level constraints. IFBENCH~\cite{ifbench} extended this with out-of-domain constraints and used programmatic verification as reinforcement learning rewards. In both cases, verification occurs \emph{post-hoc}: code checks whether model outputs satisfy prescribed format rules. Our approach differs fundamentally: we apply programmatic verification \emph{during benchmark construction}, not evaluation. Rather than checking output formats, we verify the \emph{semantic correctness} of generated conditions by executing predicates against extracted visual facts. This ensures benchmark data is logically sound by design, eliminating contradictions that arise when LLMs directly generate complex instructions. In short, prior work uses code to judge outputs; we use code to guarantee data quality.

\paragraph{Compositional and Logical Visual Reasoning.}
Recent advancements evaluate MLLMs beyond basic perception by targeting compositional relations, spatial intelligence, and logic~\cite{zerroug2022benchmark, zhang2019raven, jiang2024marvel, yangspace, yang2025mmsi}. Frameworks such as VisuLogic~\cite{xu2025visulogic}, VER-Bench~\cite{qiang2025ver}, and LogicVista~\cite{xiao2024logicvista} challenge models with visual-centric puzzles that demand fine-grained evidence extraction to preclude text-only shortcuts. Concurrently, multi-step capabilities and rigorous analytical deductions are assessed through sequential reasoning tasks~\cite{lu2024mathvista, masry2022chartqa, zhang2024mathverse, qian2025prism}. Our approach differs in structure: while existing frameworks predominantly evaluate single-layer compositions, isolated visual relations, or sequential reasoning without verified branching, \textbf{{\color{cond} MM-CondChain}} targets visually grounded deep compositional reasoning under multi-layer control flow. At each step, the model must verify a compositional visual condition, and the outcome of one step determines the next reasoning path.

\paragraph{Complex Visual Instruction Following.}
The evaluation of instruction following has recently transitioned from purely textual constraints to multi-modal and cross-contextual environments. Benchmarks like MIA-Bench~\cite{qian2024mia}, VC-IFEval~\cite{he2026empowering}, and MC-Bench~\cite{xu2025mcbench} test the strict adherence of MLLMs to layered, visual-centric directives. To navigate these complex tasks, models increasingly leverage structured inference paradigms such as Visual Chain-of-Thought (VCoT), Visual-Interleaved CoT, and step-by-step curriculum learning~\cite{chenmint, thawakar2025llamav, shao2024vcot, wu2025vic}. Our approach differs structurally: prior visual instruction datasets usually present flat, additive constraints, where missing one visual detail mainly reduces an overall compliance score. In contrast, \textbf{{\color{cond} MM-CondChain}} organizes instructions as multi-layer chains of compositional visual conditions, so that failing one condition changes the downstream execution path. Moreover, VPIR allows us to pair each verified chain with a minimally perturbed counterfactual, producing mechanically verified hard negatives that are nearly identical in wording but differ in execution outcome.
\section{VPIR-based Agentic Benchmark Construction}

\subsection{Overview}

\begin{figure}[t]
  \centering
  \includegraphics[width=\textwidth]{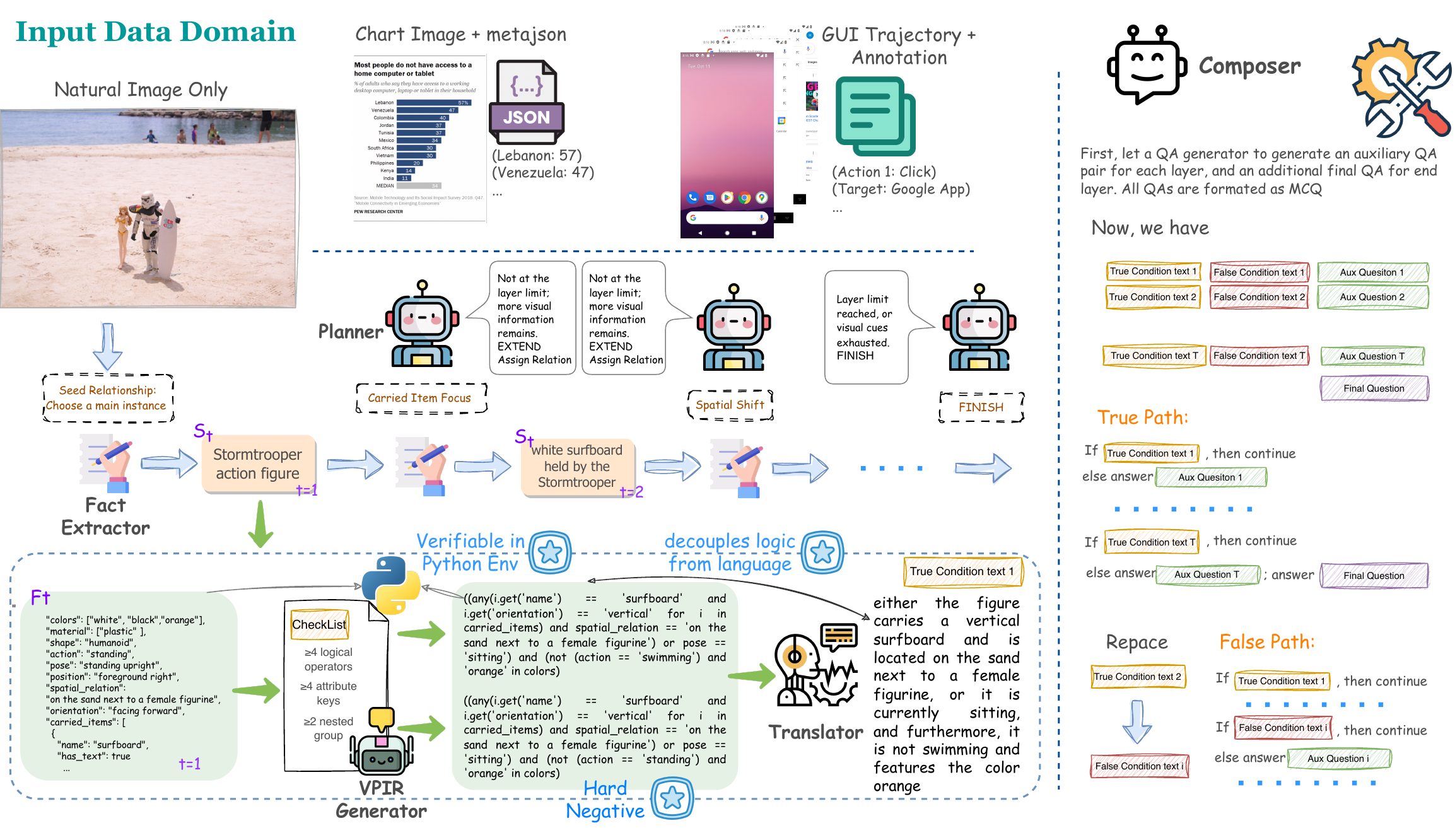}
  \caption{\textbf{Overview of the MM-CondChain agentic synthesis pipeline.}
  Given a multimodal input, the Planner iteratively extends a conditional
  chain: at each layer, structured facts are extracted, a VPIR predicate
  pair is generated and verified via code execution, and the logic is
  rendered into natural language. The Composer then compiles the verified
  chain into paired True-path and False-path instances for evaluation.}
  \label{fig:pipeline}
\end{figure}

Directly prompting an MLLM agent to generate long, multi-layer compositional reasoning chains often leads to logical inconsistencies and unverifiable claims. To address this, we propose a VPIR-based agentic benchmark construction pipeline that \emph{decouples logical construction from language rendering}. The core idea is to first construct a Verifiable Programmatic Intermediate Representation (VPIR), which is executable Python-like predicates whose truth values can be mechanically verified against visual facts. We then render the verified logic into natural language. Figure~\ref{fig:pipeline} illustrates the overall pipeline.

Given a multimodal input (\textit{e.g.}, a natural image, a chart, or a GUI trajectory), the pipeline iteratively builds a multi-layer reasoning chain. At each layer, it selects a visually grounded subject, extracts structured facts, generates an executable VPIR predicate, and renders the verified predicate into natural language (\textbf{Sec.~\ref{VPIR}}). Each layer must pass verification before the chain can extend further.

To coordinate chain construction, a Planner (\textbf{Sec.~\ref{Planner}}) decides whether to extend, terminate, or rollback the chain, working together with a Verifier (\textbf{Sec.~\ref{Verifier}}) that performs quality control. Finally, a Composer (\textbf{Sec.~\ref{Compose}}) compiles each verified chain into paired benchmark instances: a True-path where all conditions hold, and a False-path where one condition is replaced by a minimally perturbed counterfactual. 
This near-isomorphic design yields hard negatives that require both precise visual grounding and deep compositional reasoning

\subsection{Layer-wise VPIR Synthesis: Facts, Strategy, and Programmatic Logic}
\label{VPIR}

We construct a deep control-flow chain iteratively, where each layer depends on the successful verification of its predecessors.
At each layer $t$, the pipeline synthesizes verifiable layer logic through a four-stage workflow: (1) selecting a relational strategy $r_t$ that constrains subject transition,
(2) extracting structured facts $F_t$ grounded in visual evidence,
(3) generating the programmatic predicate pair $(p_t, \tilde{p}_t)$,
and (4) rendering executable logic into natural language. This decoupling of \emph{logic formation} from \emph{language rendering} ensures that truth values are mechanically computable before any linguistic expression.

\subsubsection{Step 1: Relational Strategy \& Subject Selection}

At each layer \(t\), we choose a relational strategy \(r_t \in \mathcal{R}\), where \(\mathcal{R}\) is a discrete taxonomy of inter-layer relations (e.g., \textit{Deepening} vs.\ \textit{Transition}).
Intuitively, \textit{Deepening} continues reasoning about the same subject by zooming into its parts or new attribute dimensions, while \textit{Transition} moves to a distinct but related entity via spatial/semantic relations.

Given the input sample \(x\) and the execution-ordered chain history \(H_{t-1}\), we instantiate \(r_t\) as a subject filter and construct a feasible set of visually grounded candidates:
\begin{equation}
\Omega_t \triangleq \Omega(x, H_{t-1}, r_t).
\end{equation}
We use \(\Omega_t\) to constrain the extractor in Step~2, which selects the subject and extracts facts jointly.
Here \(H_{t-1}\) summarizes previous layers in execution order, including their selected subjects and verification outcomes, since the control flow is evaluated sequentially along the chain.

\subsubsection{Step 2: Structured Fact Extraction}

To prevent hallucination during logic synthesis, the pipeline grounds generation in a structured, domain-agnostic factual representation. Conditioned on $r_t$ (and thus \(\Omega_t\)) and history $H_{t-1}$, the extractor jointly selects a grounded subject \(S_t \in \Omega_t\) and produces the subject--fact pair:
\begin{equation}
(S_t, F_t) = \mathcal{E}(x, r_t, H_{t-1}).
\end{equation}
For the seed layer ($t=1$), $H_0 = \emptyset$ and $r_1$ is a foundational seed strategy.
The extracted facts $F_t$ constitute a typed key-value mapping $\{(k, v_k)\}$\footnote{``Typed'' means values in $F_t$ use JSON-compatible types (e.g., \texttt{str}/\texttt{int}/\texttt{float}/\texttt{bool}, \texttt{list}/\texttt{dict}) and are exposed as variables for VPIR execution. VPIR only permits whitelisted primitives (e.g., \texttt{len}, \texttt{any/all}, \texttt{min/max/sum}) on these types, ensuring deterministic verifiability.}, where each key $k$ denotes a visual attribute dimension (e.g., \texttt{color}, \texttt{spatial\_relation}, \texttt{count}, \texttt{gui\_state}) and $v_k$ is a typed observation (e.g., \textit{red}, \textit{left-of}, \textit{50}, \textit{list-layout}).

We enforce two critical design principles:

\begin{itemize}
    \item \textbf{Object-Centric Grounding:} The subject $S_t$ must be uniquely localizable in the visual input, ensuring conditions are rooted in visual evidence.
    \item \textbf{Structure-First Representation:} By representing \(F_t\) as a JSON dictionary (rather than free-form text), we define a programmatic namespace \(\mathcal{V}_t \triangleq \mathrm{keys}(F_t)\), enabling mechanical verification via executable semantics.
\end{itemize}

\subsubsection{Step 3: VPIR Generation}

With the fact space $F_t$ and variable namespace $\mathcal{V}_t$ established, the pipeline synthesizes the Verifiable Programmatic Intermediate Representation (VPIR). We define the VPIR at layer $t$ as a pair of executable predicate programs: the true-logic $p_t$ and the counterfactual false-logic $\tilde{p}_t$.

To formally verify these predicates, we evaluate VPIR in a sandboxed execution environment \(\text{Env}(F_t)\). This environment exposes only whitelisted built-in operators \(\mathbb{B}\) (e.g., \texttt{len}, \texttt{set}, \texttt{all}, \texttt{any}) and binds each fact key \(k \in \mathcal{V}_t\) to its extracted value \(F_t[k]\). The semantics of a VPIR predicate is then defined by its deterministic boolean output:
\begin{equation}
\llbracket p \rrbracket(F_t) \triangleq \text{Exec}(p; \text{Env}(F_t)) \in \{0, 1\}.
\end{equation}

This programmatic formulation \textbf{guarantees absolute verifiability}, the generated predicates are accepted only by mechanical execution against $F_t$:
\begin{equation}
\llbracket p_t \rrbracket(F_t) = 1, \quad \llbracket \tilde{p}_t \rrbracket(F_t) = 0.
\end{equation}

Furthermore, through prompt-based constraints, we \textbf{encourage} (i) \textbf{non-trivial predicate complexity} (e.g., multi-clause boolean compositions with nested structure and multiple fact keys) and (ii) \textbf{minimal counterfactual perturbations} in \(\tilde{p}_t\) relative to \(p_t\), so that True/False instances remain nearly isomorphic in surface form and cannot be distinguished by shallow textual cues.

\subsubsection{Step 4: Logic Rendering}

Once the VPIR pair $(p_t, \tilde{p}_t)$ passes programmatic verification, an LLM-based Translator renders the executable logic into natural language: a true condition text $c_t$ and a counterfactual condition text $\tilde{c}_t$ (rendered from $\tilde{p}_t$). Here $\tilde{c}_t$ is retained for downstream paired-path compilation (Sec.~\ref{Compose}), where it will be substituted at a single layer to trigger early termination in the False-path instance.

Crucially, \textit{truth values are anchored in code execution; language is merely a surface rendering for evaluation}. We then apply expression-level verification (Sec.~\ref{Verifier}) to ensure the rendering is fluent, unambiguous, and faithful to the verified VPIR semantics.

\paragraph{Tiny Example.}
Consider a red car parked left of a blue truck. At layer $t$, the Planner selects $r_t = \textit{Transition}$; the extractor produces $S_t = \text{``the car''}$ with $F_t = \{\texttt{color}: \text{``red''}, \texttt{position}: \text{``left''}\}$; the pipeline generates $p_t$: \texttt{color == "red" and position == "left"} and its minimal perturbation $\tilde{p}_t$: \texttt{color == "blue" and ...}; finally, the Translator renders $c_t = \text{``the car is red and on the left''}$. Mechanical execution confirms $\llbracket p_t \rrbracket = 1$, $\llbracket \tilde{p}_t \rrbracket = 0$.


\subsection{Dedicated Verifier}
\label{Verifier}

We employ a dedicated MLLM-based Verifier for centralized quality control throughout chain construction. 

At layer \(t\), a candidate is a bundle \(\mathcal{B}_t=(S_t,F_t,p_t,\tilde p_t,c_t,\tilde c_t)\).
The Verifier returns a structured verdict \(\mathbf{v}=\{\texttt{passed},\texttt{reasons},\texttt{fix\_hint}\}\).
Verification proceeds in two stages:

\paragraph{Stage I: Fact and Subject Verification.}
Stage~I validates the grounded materials \((S_t, F_t)\) before any language rendering occurs. It checks:
\begin{itemize}
    \item \textbf{Visual Grounded}: \(S_t\) must be uniquely localizable in the input \(x\);
    \item \textbf{Non-Repetition}: the subject and extracted facts must not duplicate those in \(H_{t-1}\);
    \item \textbf{Relational Compliance}: the selection must satisfy the chosen strategy \(r_t\);
    \item \textbf{Schema \& Consistency}: \(F_t\) must conform to the domain schema with coherent cross-attribute values.
\end{itemize}

\paragraph{Stage II: Language Realization Verification.}
Stage~II validates the rendered natural-language conditions \((c_t, \tilde{c}_t)\) against the verified VPIR predicates \((p_t, \tilde{p}_t)\). It checks:
\begin{itemize}
    \item \textbf{Semantic Fidelity}: the natural language must preserve the VPIR logic without residual code artifacts;
    \item \textbf{Unambiguous Reference}: each clause must explicitly name its subject, avoiding coreference ambiguity;
    \item \textbf{Counterfactual Quality}: \(\tilde{c}_t\) must faithfully reflect \(\tilde{p}_t\) while remaining minimally perturbed from \(c_t\).
\end{itemize}

\paragraph{Feedback-Driven Regeneration.}
Verification is stage-aware: failures in Stage~I trigger regeneration of \((S_t,F_t)\), while failures in Stage~II retain the verified \((S_t,F_t,p_t,\tilde p_t)\) and only re-render \((c_t,\tilde c_t)\).

\vspace{-2mm}
\subsection{Planner: Verification-Aware Chain Control}
\label{Planner}

We introduce a verification-aware Planner that governs chain-level control flow. This dynamic interplay between the MLLM-based Planner and the Verifier constitutes the \emph{agentic} core of our pipeline: the Planner proposes actions, the Verifier provides feedback, and the Planner adapts accordingly.

At each layer \(t\), the Planner outputs a decision \((a_t, r_t) = \pi(H_{t-1})\), where \(a_t\) is an action and \(r_t \in \mathcal{R}\) is a relational strategy (Sec.~\ref{VPIR}, Step~1).
The action space consists of three options:
\begin{itemize}
    \item \texttt{EXTEND}: synthesize a new layer under the proposed strategy \(r_t\);
    \item \texttt{FINISH}: terminate the chain and proceed to composition;
    \item \texttt{ROLLBACK}: discard the most recent non-seed layer and resume from a verified prefix.
\end{itemize}

\subsubsection{Hybrid Depth Control}
The Planner combines hard-coded rules with an MLLM-driven policy. Given a target depth interval \([d_{\min}, d_{\max}]\):
\begin{itemize}
    \item If \(\mathrm{depth}(H_{t-1}) < d_{\min}\): force \(a_t = \texttt{EXTEND}\);
    \item If \(\mathrm{depth}(H_{t-1}) \ge d_{\max}\): force \(a_t = \texttt{FINISH}\);
    \item Otherwise: delegate to \(a_t = \pi_{\text{MLLM}}(H_{t-1})\), an MLLM-based policy that decides based on chain coherence and remaining synthesis potential.
\end{itemize}

\vspace{-2mm}
\subsubsection{Verification-Aware Backtracking}
The Planner is tightly coupled with the Verifier (Sec.~\ref{Verifier}). When repeated verification failures occur at the current frontier~(e.g., persistent subject repetition or unsatisfiable relational constraints), the Planner triggers \texttt{ROLLBACK}, pruning the failing layer and resuming synthesis from the last verified prefix. This feedback loop prevents the pipeline from getting stuck in unrecoverable states.

Once the Planner emits \texttt{FINISH}, the chain is finalized and forwarded to the Composer (Sec.~\ref{Compose}).


\subsection{Composition: Paired-Path Instruction Compilation}
\label{Compose}

After the Planner emits \texttt{FINISH}, we obtain a verified control-flow skeleton comprising \(T\) layers, where each layer \(t\) provides a grounded subject \(S_t\) and its true/counterfactual conditions \((c_t, \tilde{c}_t)\).

Since the control flow may terminate at any layer, we attach a question to each possible exit point: a \emph{final question} \(q^{\text{fin}}\) for the terminal layer, and an \emph{auxiliary question} \(q_t^{\text{aux}}\) for each intermediate layer. All questions are \textbf{multiple-choice} with deterministic answers. Unlike prior complex-instruction benchmarks that depend on LLM-as-judge for open-ended evaluation
\cite{zhang2025inverse,yang2025mars,deshpande2025multichallenge,zou2025eifbench,yao2023collie,wen2024benchmarking,qian2024mia}, our design enables fully reproducible and objective scoring. The \textbf{Composer} compiles this skeleton into evaluation-ready instances through two steps.

\paragraph{Step 1: Subject De-leakage.}
A subject description may inadvertently reveal its associated condition. For example, if a condition tests ``whether the car is red,'' describing the subject as ``the red car'' would leak the answer. To prevent this, an MLLM-based rewriter rephrases each \(S_t\) into a \emph{safe subject} \(\bar{S}_t\) by removing condition-revealing attributes and substituting alternative visually grounded descriptors (e.g., spatial location) when needed.
The core constraint is that \(\bar{S}_t\) must remain uniquely referential, i.e., it should still unambiguously identify the same target object \(S_t\) in the visual input.

\paragraph{Step 2: Paired-Path Instantiation.}
From each skeleton, we compile two nearly isomorphic evaluation instances:
\begin{itemize}
    \item \textbf{True-path}: All conditions \(\{c_t\}_{t=1}^{T}\) hold, so the control flow reaches the terminal layer and the correct answer corresponds to \(q^{\text{fin}}\).
    \item \textbf{False-path}: We uniformly sample a divergence layer \(j \in \{1, \dots, T{-}1\}\) and swap \(c_j \leftarrow \tilde{c}_j\). Since \(\llbracket \tilde{p}_j \rrbracket(F_j) = 0\), the flow terminates early at layer \(j\), and the correct answer becomes \(q_j^{\text{aux}}\).
\end{itemize}
Finally, we merge each \((\bar{S}_t, c_t)\) into a fluent natural-language if-clause to produce the final nested instruction. This paired compilation creates a \emph{hard negative}: the two paths share identical structure and nearly identical wording, differing only in a single subtly perturbed condition hidden among multiple true ones. Distinguishing them thus requires fine-grained reasoning over each condition rather than superficial pattern matching.


\subsection{Domain-Specific Instantiation}
\label{Domain}

The VPIR synthesis pipeline is domain-agnostic at its core; domain-specific adaptations are confined to input preprocessing and fact extraction. We instantiate the framework across three visual domains (natural images, data charts, and GUI trajectories), each requiring different input normalization before entering the unified engine (Table~\ref{tab:domain}).

\begin{table}[t]
\centering
\caption{\textbf{Domain-specific adaptations within the unified VPIR framework.}}
\setlength{\tabcolsep}{10pt}
\resizebox{\columnwidth}{!}{%
\begin{tabular}{l|ccc}
\toprule
\textbf{Aspect} & \textbf{Natural} & \textbf{Chart} & \textbf{GUI} \\
\midrule
Input & Single image & Image + metadata & Image seq.\ + annotation \\
Preprocessing & None & CSV align + LLM repair & Completeness + CoAT parse \\
Fact Focus & Visual attributes & Numerical stats & Temporal actions \\
\bottomrule
\end{tabular}%
}

\label{tab:domain}
\end{table}

\paragraph{Natural Images.}
No preprocessing is required; the MLLM directly extracts open-schema visual attributes (e.g., color, spatial relations) from the raw image.

\paragraph{Charts.}
ChartQA annotations often exhibit x/y length mismatches and zero-placeholder artifacts (missing data points marked by null bounding boxes). We apply deterministic CSV alignment to fix length inconsistencies, and LLM-based value extraction to repair missing entries, producing clean \texttt{meta\_json} before invoking the engine.

\paragraph{GUI Trajectories.}
We verify trajectory completeness (ensuring screenshot count matches annotation length), parse CoAT action descriptions into structured fields per step (action type, target element, location, etc.), and pass the multi-image sequence to the engine.

\medskip
\noindent
Crucially, the core components (VPIR predicate generation, two-stage verification, and Planner backtracking) remain \emph{entirely domain-agnostic}. Domain-specific code is isolated to input adapters, fact builders, and strategy registries. This demonstrates that the VPIR abstraction generalizes across visual modalities, from unconstrained natural scenes to structured data visualizations and interactive interface trajectories. Full preprocessing details are provided in Appendix.
\section{Evaluation}

\subsection{Evaluation setup}

\begin{figure*}[t] 
\centering

\begin{subfigure}[t]{0.5\textwidth}
  \centering
  \includegraphics[width=\linewidth]{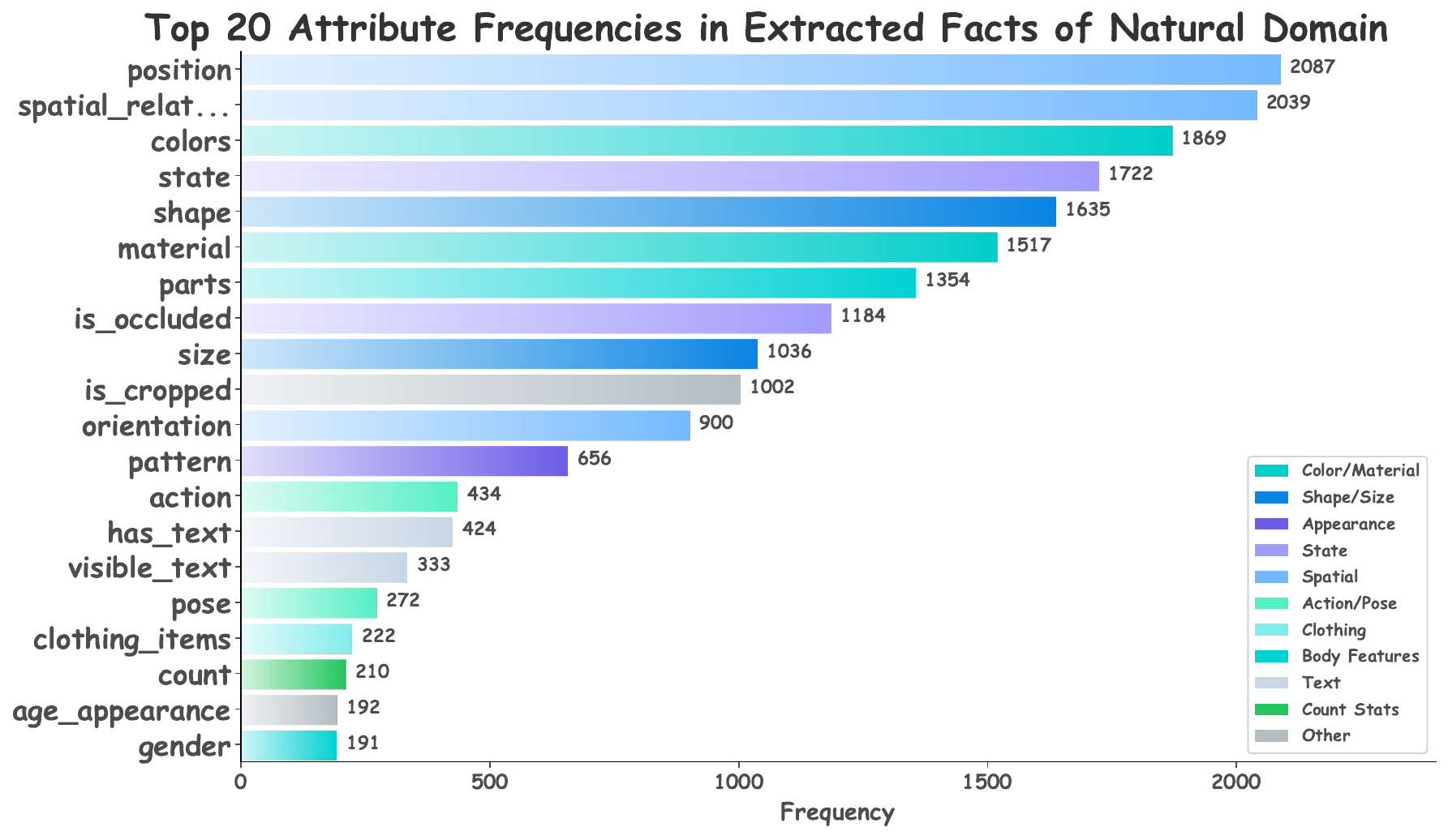}
  \caption{}
\end{subfigure}\hfill
\begin{subfigure}[t]{0.5\textwidth}
  \centering
  \includegraphics[width=\linewidth]{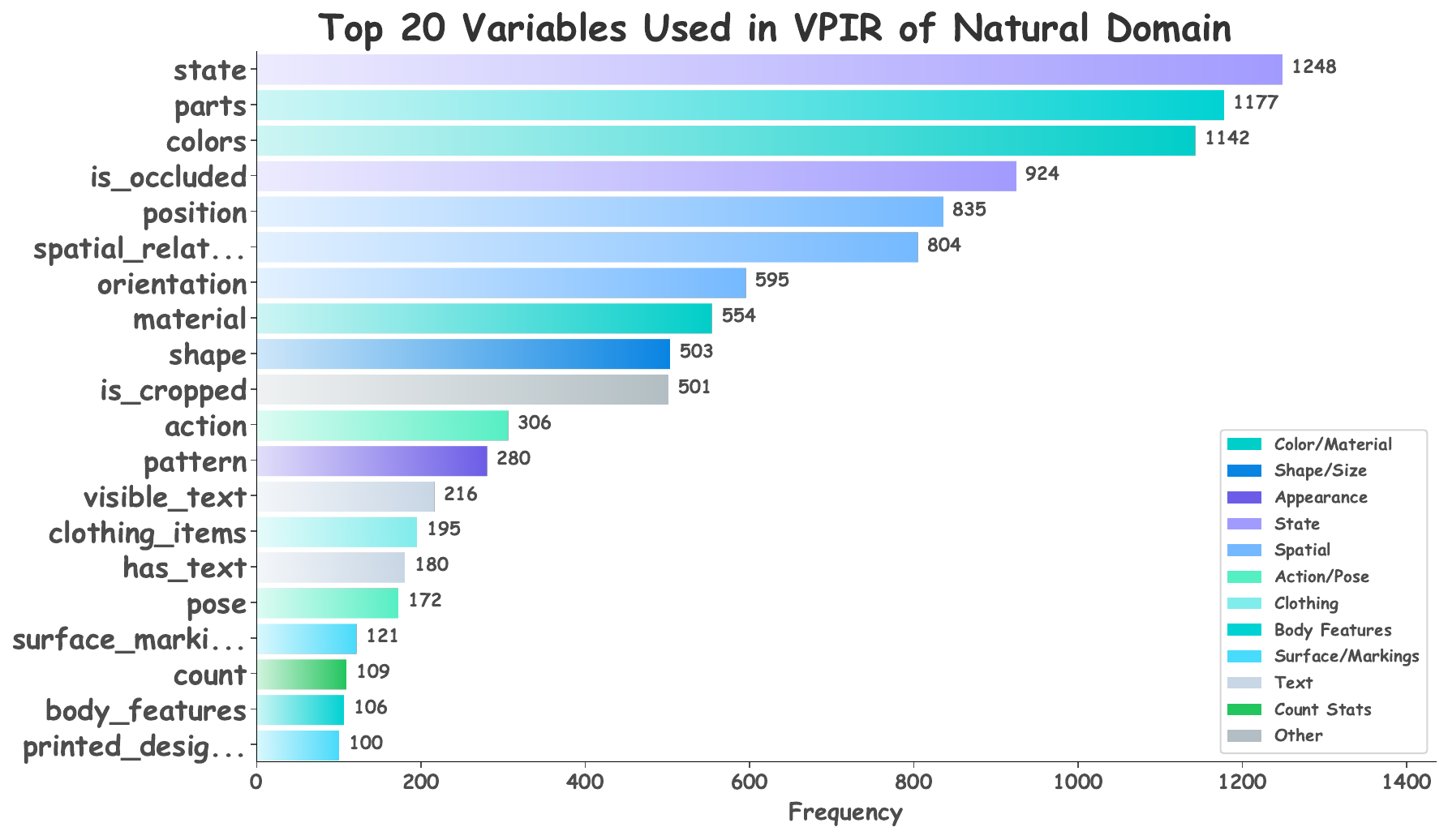}
  \caption{}
\end{subfigure}\hfill

\vspace{2mm}

\begin{subfigure}[t]{0.5\textwidth}
  \centering
  \includegraphics[width=\linewidth]{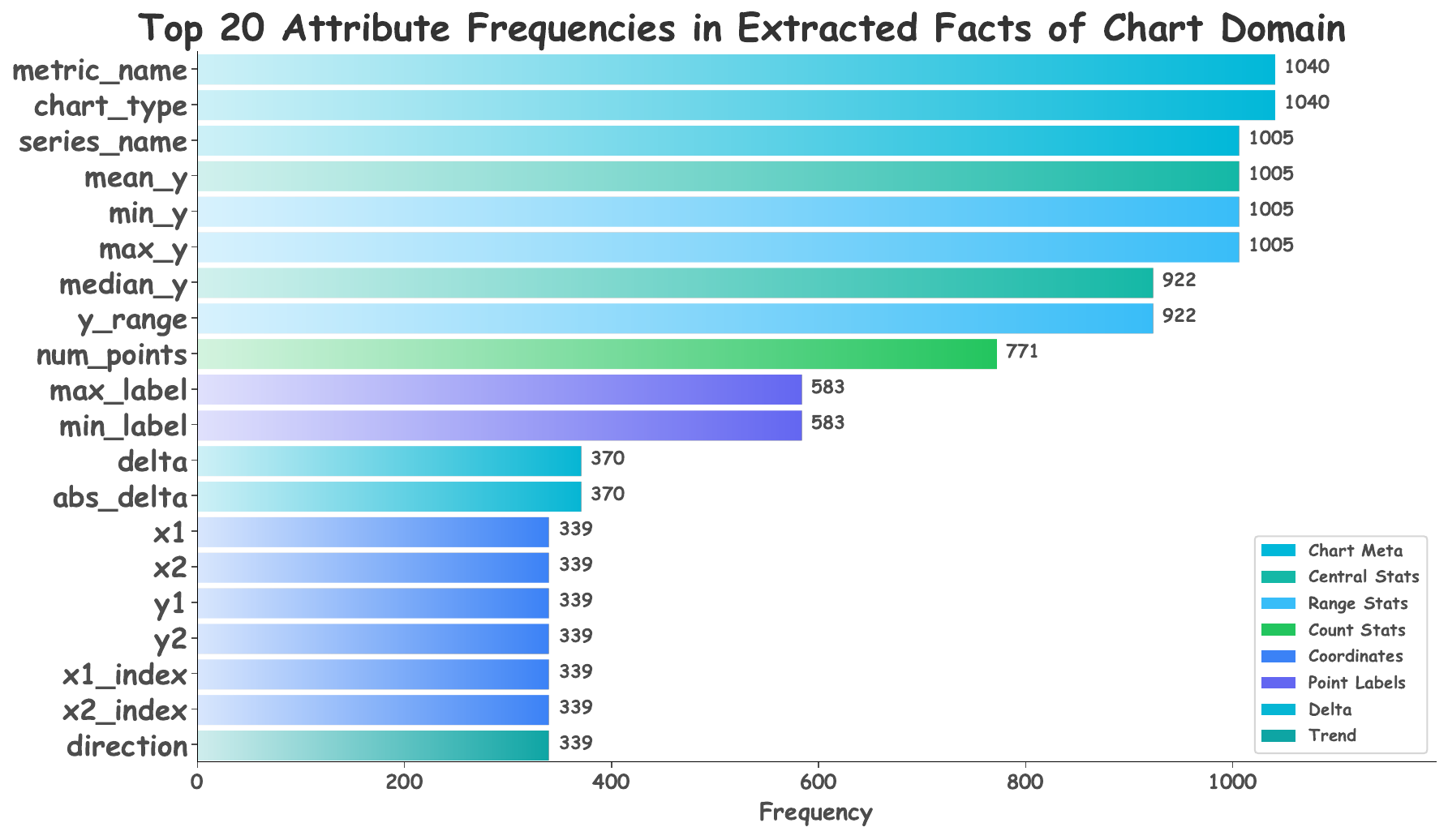}
  \caption{}
\end{subfigure}\hfill
\begin{subfigure}[t]{0.5\textwidth}
  \centering
  \includegraphics[width=\linewidth]{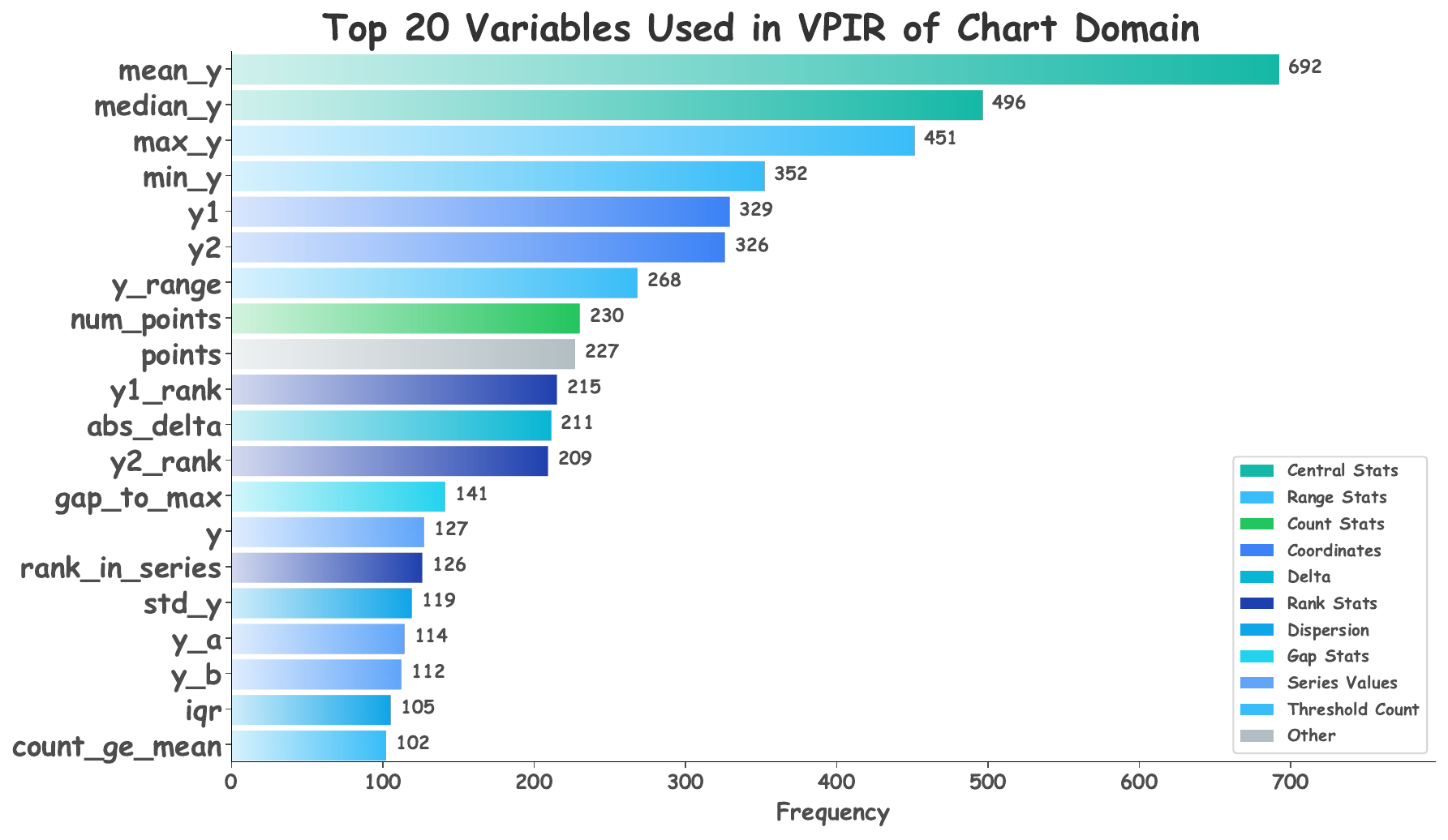}
  \caption{}
\end{subfigure}\hfill

\vspace{2mm}

\begin{subfigure}[t]{0.5\textwidth}
  \centering
  \includegraphics[width=\linewidth]{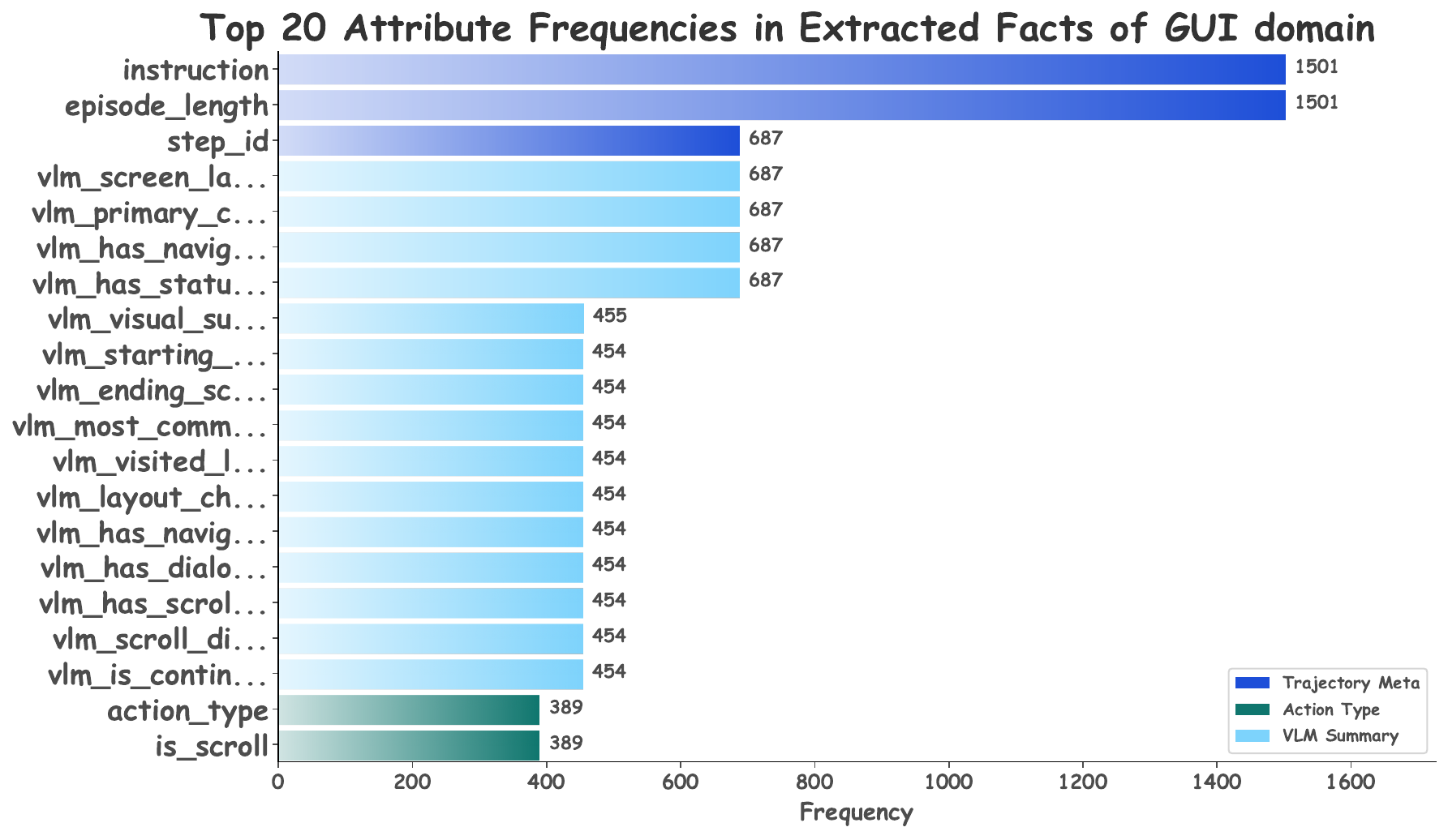}
  \caption{}
\end{subfigure}\hfill
\begin{subfigure}[t]{0.5\textwidth}
  \centering
  \includegraphics[width=\linewidth]{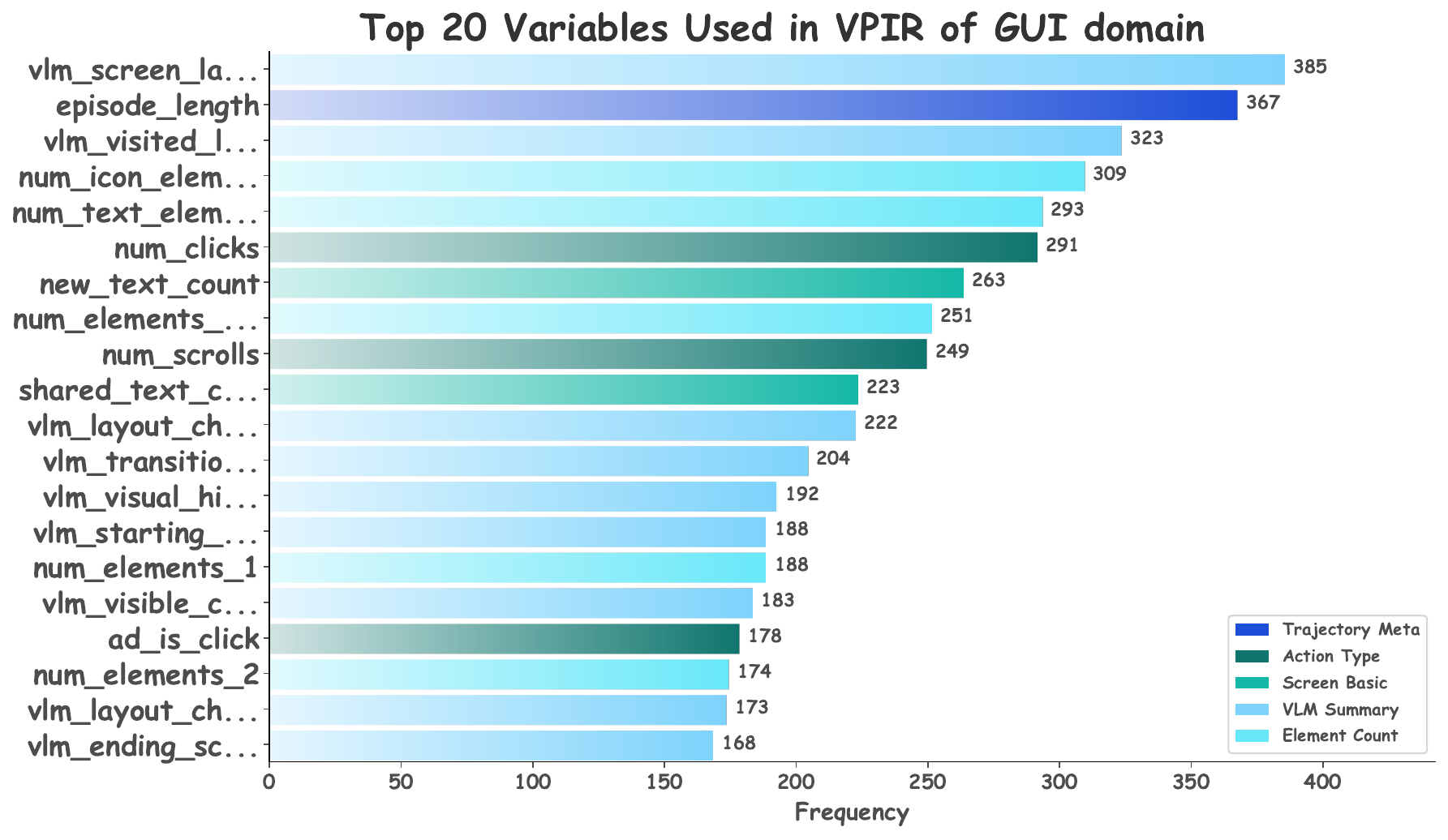}
  \caption{}
\end{subfigure}\hfill

\caption{\textbf{Top attribute frequencies in extracted facts and VPIR variables across domains.}
(a,c,e) show the top 20 attributes in extracted facts for the Natural, Chart, and GUI domains, respectively; (b,d,f) show the top 20 variables used in VPIR predicates for the corresponding domains.}
\label{fig:grid}
\end{figure*}

\begin{figure}[t]
  \centering
  \includegraphics[width=\textwidth]{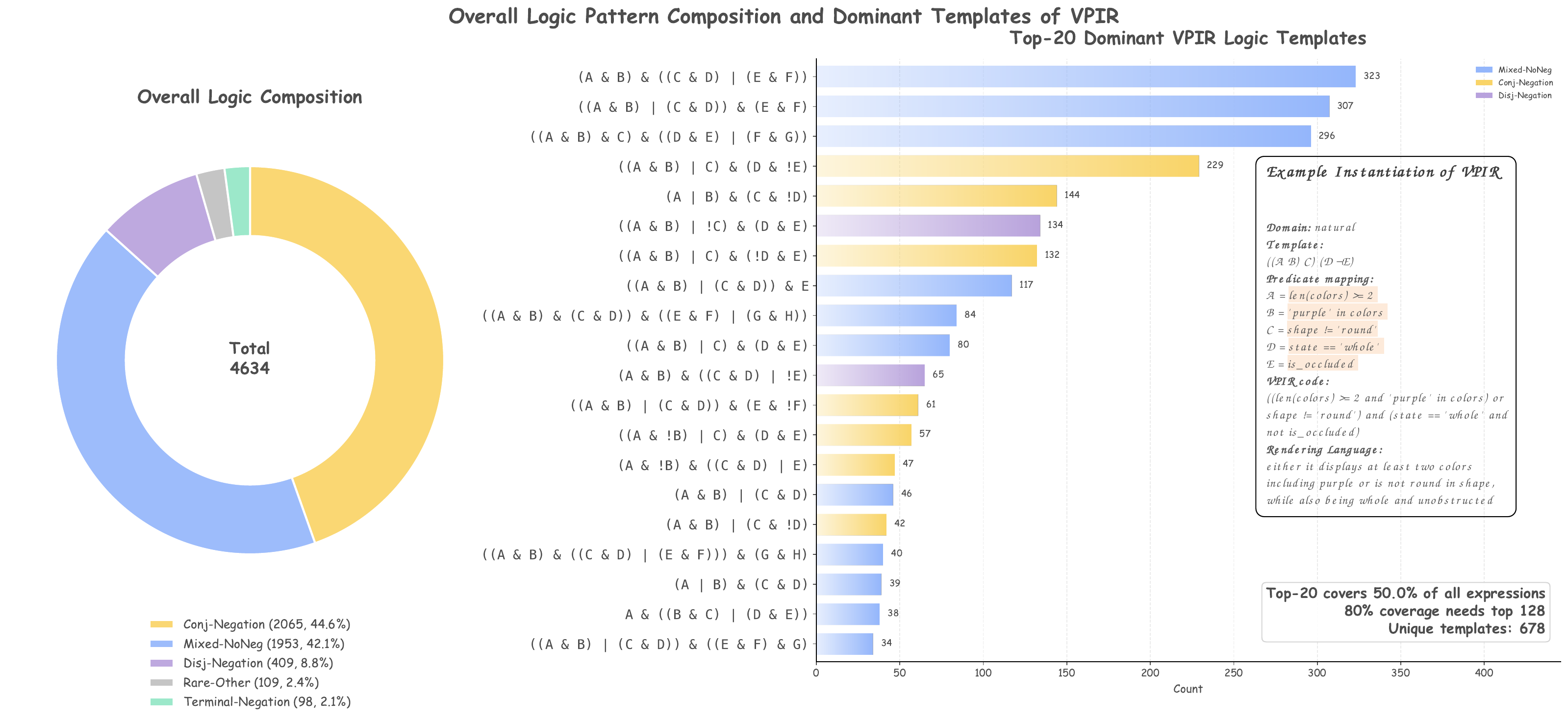}
  \caption{\textbf{Logic pattern composition of VPIR expressions.}
Left: overall distribution of high-level VPIR logic families.
Middle: top-20 dominant concrete VPIR templates.
Right: an example showing how a VPIR template is instantiated into executable predicates and natural-language conditions.}
  \label{fig:logical_pattern}
\end{figure}

\paragraph{Data Statistics.}
We construct \textbf{{\color{cond} MM-CondChain}} from three visual domains using 
publicly available datasets.
The Natural domain comprises 398 images drawn from SAM~\cite{kirillov2023segment} (204) 
and GQA~\cite{hudson2019gqa} (194).
The Chart domain includes 200 chart images from ChartQA~\cite{masry2022chartqa}, 
spanning bar, line, and pie charts with structured numerical annotations.
The GUI domain contains 377 interaction trajectories (3,421 screenshots 
in total, averaging 9.07 frames per trajectory) sourced from 
AITZ~\cite{aitz}, which provides fine-grained reasoning annotations 
over AITW~\cite{rawles2023androidinthewild}.
This results in 975 evaluation samples in total, each containing a paired True-path and False-path instance.

\paragraph{Extracted Facts and VPIR Variable Statistics.}
Figure~\ref{fig:grid} shows the attribute distributions in extracted facts and the variables used in VPIR predicates across domains.
We observe clear domain-specific patterns: Natural instances mainly rely on object attributes and spatial relations, Chart instances concentrate on numerical and structural statistics, and GUI instances emphasize action, state, and trajectory-level metadata.
We also find that the VPIR variable distributions do not simply mirror the full extracted fact distributions.
Instead, VPIR selectively reuses subsets of extracted attributes to compose executable predicates, indicating that benchmark difficulty is driven by structured compositional reasoning over grounded visual facts rather than by raw attribute frequency alone.

\paragraph{Logical Pattern Statistics.}
Figure~\ref{fig:logical_pattern} shows that VPIR expressions in \textbf{{\color{cond} MM-CondChain}} exhibit substantial structural diversity.
Although several pattern families appear more frequently than others, the benchmark is not dominated by one or two simple templates: the top-20 templates cover only 50.07\% of all expressions, and 128 unique templates are needed to reach 80\% coverage.
This indicates that the benchmark contains a broad range of compositional logic structures rather than a small set of repeated forms.
Moreover, the dominant templates themselves are already structurally complex.
As illustrated by the example on the right, a single VPIR template can involve multiple predicates, nested logical operators, executable program form, and its corresponding natural-language rendering.
As a result, correctly solving these instances requires not only visual grounding of the relevant objects, attributes, and relations, but also compositional reasoning over how these visual factors jointly determine whether the condition holds.

\paragraph{Benchmark Generation.}
We employ \texttt{Gemini-3-Pro}~\cite{deepmind_gemini3pro_model_card}, currently among the strongest MLLMs in comprehensive reasoning capabilities, to instantiate all MLLM and LLM agents in our synthesis pipeline, including the Planner, Verifier, Fact Extractor, and Translator.

\paragraph{Evaluated Models.}
We evaluate a range of MLLMs spanning both open-source and proprietary families.
Open-source models include the Qwen3-VL series~\cite{bai2025qwen3vl}, Qwen3.5 series~\cite{qwen3.5}, GLM-4.6V series~\cite{vteam2025glm45vglm41vthinkingversatilemultimodal}, Kimi-K2.5~\cite{team2026kimi}, InternVL3 series~\cite{zhu2025internvl3} and InternVL3.5-8B~\cite{wang2025internvl35}.
Proprietary models include GPT-4o-1120~\cite{achiam2023gpt4}, GPT-5-0807~\cite{openai_gpt5_system_card}, Gemini-2.5-Flash~\cite{deepmind_gemini25flash_modelcard_pdf}, Gemini-2.5-Pro~\cite{deepmind_gemini25pro_modelcard_pdf}, Gemini-3-Flash~\cite{deepmind_gemini3flash_model_card}, Gemini-3-Pro~\cite{deepmind_gemini3pro_model_card}, Qwen3-VL-Flash, and Qwen3-VL-Plus.

\paragraph{Evaluation Metrics.}
We report three metrics for each domain:
(1)~\textbf{True-path Accuracy}: the percentage of True-path instances where the model correctly follows all conditions and selects the answer corresponding to the final question $q^{\mathrm{fin}}$;
(2)~\textbf{False-path Accuracy}: the percentage of False-path instances where the model correctly identifies the early-termination point and selects the auxiliary answer $q_j^{\mathrm{aux}}$;
(3)~\textbf{Path F1}: the harmonic mean of True-path and False-path accuracy, measuring balanced performance across both paths.
We also report \textbf{Avg(F1)}, the arithmetic mean of Path F1 across three domains, as the overall score.

\paragraph{Implementation Details of Evaluation.}
All models are evaluated in a zero-shot setting using each provider's default API parameters (temperature, max tokens, etc.).
Each instance is presented as a multiple-choice question with a specified output format.
Answers are extracted by prioritizing the last \texttt{\textbackslash boxed\{\ldots\}} match, with fallback to standalone option patterns; unparseable outputs are marked incorrect.

\subsection{Main Results}

\definecolor{FOneNat}{HTML}{EAF2FF}   
\definecolor{FOneChart}{HTML}{EAF7EA} 
\definecolor{FOneGUI}{HTML}{FFF1E6}   
\definecolor{FOneAvg}{HTML}{F2F2F2}   
\definecolor{DarkGray}{HTML}{555555}  

\begin{table*}[t]
\centering
\caption{\textbf{Main results on MM-CondChain across domains.} All numbers are percentages (\%). Path F1 is the harmonic mean of True- and False-path accuracy. Avg(F1) is the mean of the three domain F1 scores. Rows are sorted by Avg(F1) in ascending order within each category.}
\label{tab:main_results_domains}
\small
\setlength{\tabcolsep}{5pt}
\resizebox{\textwidth}{!}{%
\begin{tabular}{l|cc>{\columncolor{FOneNat}}c|cc>{\columncolor{FOneChart}}c|cc>{\columncolor{FOneGUI}}c|>{\columncolor{FOneAvg}}c}
\toprule
& \multicolumn{3}{c|}{\textbf{Natural}} & \multicolumn{3}{c|}{\textbf{Chart}} & \multicolumn{3}{c|}{\textbf{GUI}} & \cellcolor{white}\textbf{Avg} \\
Model & True & False & \cellcolor{white}F1 & True & False & \cellcolor{white}F1 & True & False & \cellcolor{white}F1 & \cellcolor{white}F1 \\
\midrule
\multicolumn{11}{c}{\textcolor{DarkGray}{\textbf{\textit{Open-Source MLLMs}}}} \\
\midrule

\texttt{Qwen3.5-0.8B}  & 33.17 & 2.26 & 4.23 &   31.50 & 3.00 & 5.48 & 33.95 & 1.86 & 3.52 & 4.41 \\
\texttt{GLM-4.6V-Flash}  & 83.92 & 9.55 & 17.14 & 81.91 & 5.53 & 10.36 & 87.53 & 0.53 & 1.05 & 9.52 \\
\texttt{InternVL3-8B}  & 65.33 & 8.29 & 14.72 & 47.50 & 8.50 & 14.42 & 63.66 & 5.31 & 9.79 & 12.98 \\
\texttt{InternVL3.5-8B}  & 82.41 & 10.30 & 18.31 & 76.00 & 19.50 & 31.04 & 82.23 & 1.33 & 2.61 & 17.32 \\
\texttt{InternVL3-14B}  & 76.38 & 13.57 & 23.04 & 43.00 & 21.00 & 28.22 & 84.62 & 2.39 & 4.64 & 18.63 \\
\texttt{Qwen3.5-4B}  & 88.92 & 15.37 & 26.20 &   86.50 & 20.00 & 32.49 & 65.78 & 7.69 & 13.77 & 24.15 \\

\texttt{Qwen3.5-35B-A3B}     & 93.43 & 11.62 & 20.66 & 88.50 & 17.00 & 28.52 & 74.27 & 14.32 & 24.02 & 24.40 \\
\texttt{Qwen3-VL-30B-A3B-Instruct}    & 27.64 & 27.14 & 27.38 & 44.00 & 35.50 & 39.30 & 73.67 &  7.98 & 14.40 & 27.03 \\
\texttt{InternVL3-38B}  & 73.62 & 20.60 & 32.20 & 31.00 & 31.50 & 31.25 & 57.03 & 12.47 & 20.46 & 27.97 \\
\texttt{Qwen3.5-9B}  & 91.69 & 13.10 & 22.92 &   86.50 & 28.50 & 42.87 & 71.62 & 11.67 & 20.07 & 28.62 \\

\texttt{Qwen3-VL-8B-Instruct}& 47.98 & 30.81 & 37.52 & 39.78 & 39.78 & 39.78 & 58.67 & 12.53 & 20.65 & 32.65 \\
\texttt{GLM-4.6V}  & 73.37 & 26.13 & 38.54 & 66.00 & 34.50 & 45.31 & 30.50 & 24.40 & 27.11 & 36.99 \\
\texttt{Qwen3-VL-8B-Thinking}  & 60.71 & 30.48 & 40.58 & 49.50 & 37.00 & 42.35 & 37.14 & 27.85 & 31.83 & 38.25 \\
\texttt{Qwen3.5-122B-A10B}   & 95.48 & 20.85 & 34.23 & 84.50 & 37.50 & 51.95 & 65.78 & 23.08 & 34.17 & 40.12 \\
\texttt{Qwen3-VL-30B-A3B-Thinking}  & 30.90 & 31.16 & 31.03 & 58.00 & 56.50 & 57.24 & 40.53 & 27.73 & 32.93 & 40.40 \\
\texttt{Kimi-K2.5}  & 75.57 & 41.06 & 53.21 &   46.00 & 52.00 & 48.82& 50.93 & 25.20 & 33.72 & 45.25 \\
\texttt{Qwen3-VL-235B-A22B-Instruct}  & 62.12 & 43.94 & 51.47 &   55.00 & 61.00 & 57.84 & 62.60 & 17.24 & 27.04 & 45.45 \\
\texttt{Qwen3.5-397B-A17B}   & 52.01 & 31.16 & 38.97 & 67.00 & 52.00 & 58.55 & 40.05 & 40.32 & 40.19 & 45.90 \\
\texttt{Qwen3-VL-235B-A22B-Thinking}  &  65.49 & 39.55 & 49.31 &   61.50 & 58.50 & 59.96 & 28.91 & 33.95 & 31.23 & 46.83 \\
\midrule
\multicolumn{11}{c}{\textcolor{DarkGray}{\textbf{\textit{Proprietary MLLMs}}}} \\
\midrule
\texttt{GPT-4o-1120}         & 83.92 & 12.81 & 22.23 & 17.00 & 18.00 & 17.49 & 63.40 & 12.20 & 20.46 & 20.06 \\
\texttt{Gemini-2.5-Flash}    & 29.40 & 48.24 & 36.53 & 35.50 & 47.00 & 40.45 &  6.90 & 44.83 & 11.95 & 29.64 \\
\texttt{Qwen3-VL-Flash}      & 61.56 & 29.65 & 40.02 & 59.50 & 47.50 & 52.83 & 58.62 & 10.61 & 17.97 & 36.94 \\
\texttt{Gemini-2.5-Pro}      & 38.94 & 55.28 & 45.70 & 55.50 & 64.50 & 59.66 & 10.34 & 54.38 & 17.38 & 40.91 \\
\texttt{Qwen3-VL-Plus}       & 67.59 & 32.16 & 43.58 & 56.00 & 54.50 & 55.24 & 34.75 & 38.20 & 36.39 & 45.07 \\
\texttt{Gemini-3-Flash}      & 54.77 & 41.46 & 47.19 & 60.50 & 63.50 & 61.96 & 36.87 & 34.75 & 35.78 & 48.31 \\
\texttt{GPT-5-0807}          & 80.65 & 33.67 & 47.51 & 63.50 & 67.50 & 65.44 & 30.77 & 49.87 & 38.06 & 50.34 \\
\texttt{Gemini-3-Pro}        & 73.87 & 44.97 & 55.91 & 70.00 & 62.50 & 66.04 & 32.63 & 45.62 & 38.05 & 53.33 \\
\bottomrule
\end{tabular}%
}

\end{table*}

\paragraph{Main Results.}
The main results are summarized in Table~\ref{tab:main_results_domains}.
Overall, current MLLMs still struggle on \textbf{{\color{cond} MM-CondChain}}.
Among all evaluated models, \texttt{Gemini-3-Pro} achieves the best overall result with 53.33 average Path F1, followed by \texttt{GPT-5-0807} at 50.34.
Even the strongest model remains only slightly above 50 F1, indicating that visually grounded deep compositional reasoning under multi-layer control flow is still highly challenging for current MLLMs.

\paragraph{True vs.\ False Paths.}
A clear pattern is that many models perform substantially better on the True-path than on the False-path.
For example, \texttt{GPT-4o-1120} scores 83.92 vs.\ 12.81 on Natural, \texttt{Qwen3.5-4B} scores 88.92 vs.\ 15.37 on Natural, and \texttt{Qwen3.5-9B} scores 91.69 vs.\ 13.10 on Natural.
This gap suggests that under complex multi-layer conditions, models tend to over-assume that the conditions hold and thus favor the ``continue'' branch.
Such a bias can be risky in real visual workflows, where failing to detect a violated condition may cause the model to proceed when it should stop, switch branches, or reject the action.

\paragraph{Model Comparisons.}
Proprietary models generally outperform open-source ones in overall performance, with \texttt{Gemini-3-Pro} and \texttt{GPT-5-0807} ranking first and second, respectively.
At the same time, open-source models remain competitive in specific settings: notably, \texttt{Qwen3.5-397B-A17B} achieves the best score on GUI (F1=40.19), surpassing all proprietary models on that domain.
We also observe that Thinking models generally outperform their Instruct counterparts, suggesting that explicit reasoning-oriented models are better suited for this complex benchmark.

\paragraph{Domain-wise Difficulty.}
We observe clear domain-dependent difficulty.
GUI is the most challenging domain overall: its best F1 is only 40.19, lower than the best results on Natural (55.91) and Chart (66.04).
This is likely because GUI instances require reasoning over multi-frame trajectories, user actions, and interface state transitions, whereas many Chart conditions reduce to deterministic numerical comparisons once the relevant values are grounded.

\subsection{Design Ablations}

\begin{table}[t]
\centering
\caption{\textbf{Effect of chain depth and predicate complexity on Path F1 (\%).}
\label{ablations}
  \textbf{Left}: Performance degrades as chain depth increases, with $\sim$30\% relative drop from D=2 to D=6.
  \textbf{Right}: Increasing intra-layer predicate complexity (\textsc{Simple} vs.\ \textsc{Complex}) causes 28--36\% degradation at fixed depth.}
\label{tab:ablation}
\small

\begin{minipage}[t]{0.52\textwidth}
  \centering
  \setlength{\tabcolsep}{6pt}
  \resizebox{\textwidth}{!}{
  \begin{tabular}{l|ccc|c}
    \toprule
    \textbf{Model} & \textbf{D=2} & \textbf{D=4} & \textbf{D=6} & \textbf{$\Delta_{2 \to 6}$} \\
    \midrule
    \texttt{Gemini-3-Flash} & 70.68 & 53.85 & 47.19 & $-$33.2\% \\
    \texttt{Qwen3-VL-Plus}  & 61.51 & 52.56 & 43.58 & $-$29.1\% \\
    \texttt{GPT-4o-1120}    & 31.39 & 27.67 & 22.23 & $-$29.2\% \\
    \bottomrule
  \end{tabular}
  }
\end{minipage}
\hfill
\begin{minipage}[t]{0.45\textwidth}
  \centering
  \setlength{\tabcolsep}{6pt}
  \resizebox{\textwidth}{!}{
  \begin{tabular}{l|cc|c}
    \toprule
    \textbf{Model} & \textsc{Simp.} & \textsc{Comp.} & \textbf{$\Delta$} \\
    \midrule
    \texttt{Gemini-3-Flash} & 65.26 & 47.19 & $-$27.7\% \\
    \texttt{Qwen3-VL-Plus}  & 62.91 & 43.58 & $-$30.7\% \\
    \texttt{GPT-4o-1120}    & 34.75 & 22.23 & $-$36.0\% \\
    \bottomrule
  \end{tabular}
  }
\end{minipage}

\end{table}

\subsubsection{Effect of Chain Depth.}

To investigate how chain depth affects model performance, we construct 
ablation instances with controlled maximum depths of 2, 4, and 6 layers on the Natural domain and evaluate three representative models.
As shown in Table~\ref{ablations} Left, all models exhibit consistent 
performance degradation as chain depth increases. 
From depth 2 to depth 6, Path F1 drops by approximately 29--33\% in 
relative terms across all tested models.
Notably, this degradation is not uniform: \texttt{Gemini-3-Flash} 
suffers the largest relative drop ($-$33.2\%), despite achieving the 
highest absolute scores, suggesting that even strong models struggle to 
maintain accuracy as the number of sequential verification steps grows.

These results confirm that tracking multi-layer conditional logic poses 
a fundamental challenge for current MLLMs.  The near-linear degradation with depth indicates that errors compound  across layers, rather than being isolated to specific conditions. This underscores the value of \textbf{{\color{cond} MM-CondChain}}'s configurable depth design for probing the limits of sequential visual reasoning.

\subsubsection{Effect of Predicate Complexity.}

Beyond chain depth, we examine how \emph{intra-layer predicate complexity} 
affects model performance.
We contrast two VPIR generation settings: \textsc{Simple} predicates 
(at most 2 logical operators, at least 2 attribute keys, no nesting 
requirement) versus \textsc{Complex} predicates (at least 4 logical 
operators, 4 attribute keys, and 2 nested groups).
Both settings share the same chain depth to isolate the effect of 
compositional logic.
As shown in Table~\ref{ablations} Right, increasing predicate 
complexity leads to substantial performance drops across all models, 
with relative degradation ranging from 27.7\% to 36.0\%.
Notably, \texttt{GPT-4o-1120} suffers the largest relative decline 
($-$36.0\%), suggesting that models with weaker baseline performance 
are disproportionately affected by compositional complexity.

These results reveal that current MLLMs struggle not only with 
\emph{sequential} reasoning across layers (as shown in the depth ablation), 
but also with \emph{compositional} reasoning within a single predicate.
The two dimensions---chain depth and predicate complexity---jointly 
define the difficulty landscape of \textbf{{\color{cond} MM-CondChain}}, enabling 
fine-grained diagnosis of model capabilities.

\subsubsection{Summary.}
The above ablations reveal two orthogonal axes of difficulty in 
\textbf{{\color{cond} MM-CondChain}}: \emph{vertical} complexity (chain depth) and 
\emph{horizontal} complexity (intra-layer predicate composition).
Increasing either dimension leads to consistent and substantial 
performance degradation across all tested models, confirming that 
both sequential reasoning and compositional reasoning remain 
fundamental bottlenecks for current MLLMs.
Crucially, these two axes are independently controllable in our 
VPIR-based synthesis pipeline, enabling fine-grained difficulty 
calibration.
This design allows \textbf{{\color{cond} MM-CondChain}} to serve not only as an 
evaluation benchmark, but also as a diagnostic tool for pinpointing 
\emph{where} and \emph{why} models fail in visually grounded 
conditional reasoning.

\section{Conclusion}

In this paper, we introduce \textbf{{\color{cond} MM-CondChain}}, a benchmark for evaluating 
visually grounded deep conditional reasoning in MLLMs.
Unlike prior benchmarks that test shallow compositions or 
independent constraints, \textbf{{\color{cond} MM-CondChain}} requires tracking 
multi-layer control flow where each decision is gated by a 
visually verifiable condition.
To enable scalable construction with guaranteed correctness, we 
proposed an agentic synthesis pipeline centered on Verifiable 
Programmatic Intermediate Representation (VPIR), which decouples 
logic formation from language rendering and produces benchmark 
instances with deterministic ground truth and near-isomorphic 
hard negatives.
Experiments across three visual domains and a range of MLLMs 
reveal that \emph{visually grounded conditional reasoning remains 
a fundamental bottleneck}: even state-of-the-art models struggle 
as chain depth or predicate complexity increases.
We believe \textbf{{\color{cond} MM-CondChain}} will serve as a valuable resource 
for diagnosing model weaknesses and driving future research toward 
more robust multimodal reasoning.

{
\small
\bibliographystyle{plainnat}
\bibliography{ref}
}
%
%

\clearpage

	
\end{document}